\pdfoutput=1

\documentclass[11pt]{article}

\usepackage{emnlp2021}

\usepackage{times}
\usepackage{latexsym}

\usepackage[T1]{fontenc}

\usepackage[utf8]{inputenc}

\usepackage{microtype}

\usepackage{xcolor,colortbl}
\usepackage{hyperref}
\usepackage{multirow}
\usepackage{tablefootnote}
\usepackage{mathtools}
\usepackage{amsfonts}
\usepackage{amsmath,amssymb}
\usepackage{hhline}
\usepackage{enumitem}
\usepackage{ltablex}

\usepackage{url}
\usepackage{tablefootnote}
\makeatletter
\newcommand{\spewfootnotes}{%
	\tfn@tablefootnoteprintout%
	\global\let\tfn@tablefootnoteprintout\relax%
	\gdef\tfn@fnt{1}%
}
\makeatother

\definecolor{darkblue}{rgb}{0, 0, 0.5}
\hypersetup{colorlinks=true,citecolor=darkblue, linkcolor=darkblue, urlcolor=darkblue}

\title{Quantifying Reproducibility in NLP and ML}

\author{Anya Belz \\
  ADAPT Centre \\
  Dublin City University, Ireland \\
  \texttt{anya.belz@adaptcentre.ie} \\}

\begin{document}
\maketitle
\begin{abstract}
Reproducibility has become an intensely debated topic in NLP and ML over recent years, but no commonly accepted way of assessing reproducibility, let alone \textit{quantifying} it, has so far emerged. The assumption has been that wider scientific reproducibility terminology and definitions are not applicable to NLP/ML, with the result that many different terms and definitions have been proposed, some diametrically opposed. In this paper, we test this assumption, by taking the standard terminology and definitions from metrology and applying them directly to NLP/ML. We find that we are able to straightforwardly derive a practical framework for assessing reproducibility which has the desirable property of yielding a quantified degree of reproducibility that is comparable across different reproduction studies.
\end{abstract}

\section{Introduction}\label{sec:intro}

Reproducibility of results is coming under increasing scrutiny in the machine learning (ML) and natural language processing (NLP) fields, against the background of a perceived reproducibility crisis in science more widely \cite{baker2016reproducibility}, and NLP/ML specifically \cite{pedersen2008empiricism,mieskes-etal-2019-community}.  There have been several workshops and checklist initiatives on the topic,\footnote{Reproducibility in ML Workshop at ICML'17, ICML'18 and ICLR'19; Reproducibility Challenge at ICLR'18, ICLR'19, NeurIPS'19, and NeurIPS'20; reproducibility track and shared task \cite{branco-etal-2020-shared} at LREC'20; reproducibility programme at NeurIPS'19 comprising a code submission policy, a reproducibility challenge for ML results, and the ML Reproducibility checklist \cite{whitaker2017}, later also adopted by EMNLP'20 and AAAI'21.} conferences are promoting reproducibility via calls, chairs' blogs,\footnote{\url{https://2020.emnlp.org/blog/2020-05-20-reproducibility}} and special themes, and the first shared tasks are being organised, including REPROLANG'20 \cite{branco-etal-2020-shared} and ReproGen'21 \cite{belz-etal-2020-reprogen}. The biggest impact so far has been that sharing of code, data and supplementary material providing increasingly detailed information about data, systems and training regimes is now expected as standard.
Yet early optimism that "[r]eproducibility would be quite easy to achieve in machine learning simply by sharing the full code used for experiments" \cite{sonnenburg2007need} is now giving way to the realisation that even with full resource sharing and original author support the same scores can often not be obtained (see Section~\ref{ssec:f1}).

Despite a growing body of research, no consensus has so far emerged about standards, terminology and definitions. Particularly for the two most frequently used terms, \textit{reproducibility} and \textit{replicability}, divergent definitions abound, variously conditioned on
same vs.\ different team, methods, artifacts, code, software, and data. E.g.\ for the ACM \cite{acm2020artifact}, \textit{results} have been \textit{reproduced} if obtained in a different study by a different team using artifacts supplied in part by the original authors, and \textit{replicated} if obtained in a different study by a different team using artifacts not supplied by the original authors.
\citet{drummond2009replicability} argues that what ML  calls \textit{reproducibility} is in fact \textit{replicability} which is the ability to re-run an experiment in exactly the same way, whereas true \textit{reproducibility} is the ability to obtain the same result by different means.
For
\citet{rougier2017sustainable}, ``\textit{[r]eproducing} the result of a computation means running the same software on the same input data 
and obtaining the same results. 
[...]. \textit{Replicating} a published result means writing and then running new software based on the description of a computational model or method provided in the original publication''.
\citet{wieling2018reproducibility} tie \textit{reproducibility} to ``the same data and methods,'' and \citet{whitaker2017}, followed by \citet{schloss2018identifying}, tie definitions of \textit{reproducibility, replicability, robustness} and \textit{generalisability} to different combinations of same vs.\ different data and code.

Underlying this diversity of definitions is the assumption that general reproducibility terminology and definitions somehow don't apply to computer science for which different terminology and definitions are needed. To the extent that terminology and definitions have been proposed for NLP/ML, these have tended to be sketchy and incomplete, e.g.\ the definitions above entirely skirt the issue of how to tell if two studies are or are not `the same' in terms of code, method, team, etc. 

In this paper we take the general definitions  (Section~\ref{sec:vim-def}) of the International Vocabulary of Metrology (VIM) \cite{jcgm2012international}, and explore how these can be mapped to the NLP/ML context and extended into a practical framework for reproducibility assessment capable of yielding quantified assessments of degree of reproducibility (Section~\ref{sec:mapping}). We start with two informal examples (Section~\ref{sec:meas-ex}), one involving weight measurements of museum artifacts, the other weighted F1 measurements of a text classifier, and conclude with a discussion of limitations and future directions (Sections~\ref{sec:disc}, \ref{sec:concl}).

\section{Example Physical and Non-physical Measurements}\label{sec:meas-ex}

\subsection{Mass of a torc}\label{ssec:torc}

\begin{figure}
\begin{tabular}{ll}
           \includegraphics[width=.275\textwidth,trim={18cm 0cm 17cm 0cm},clip]{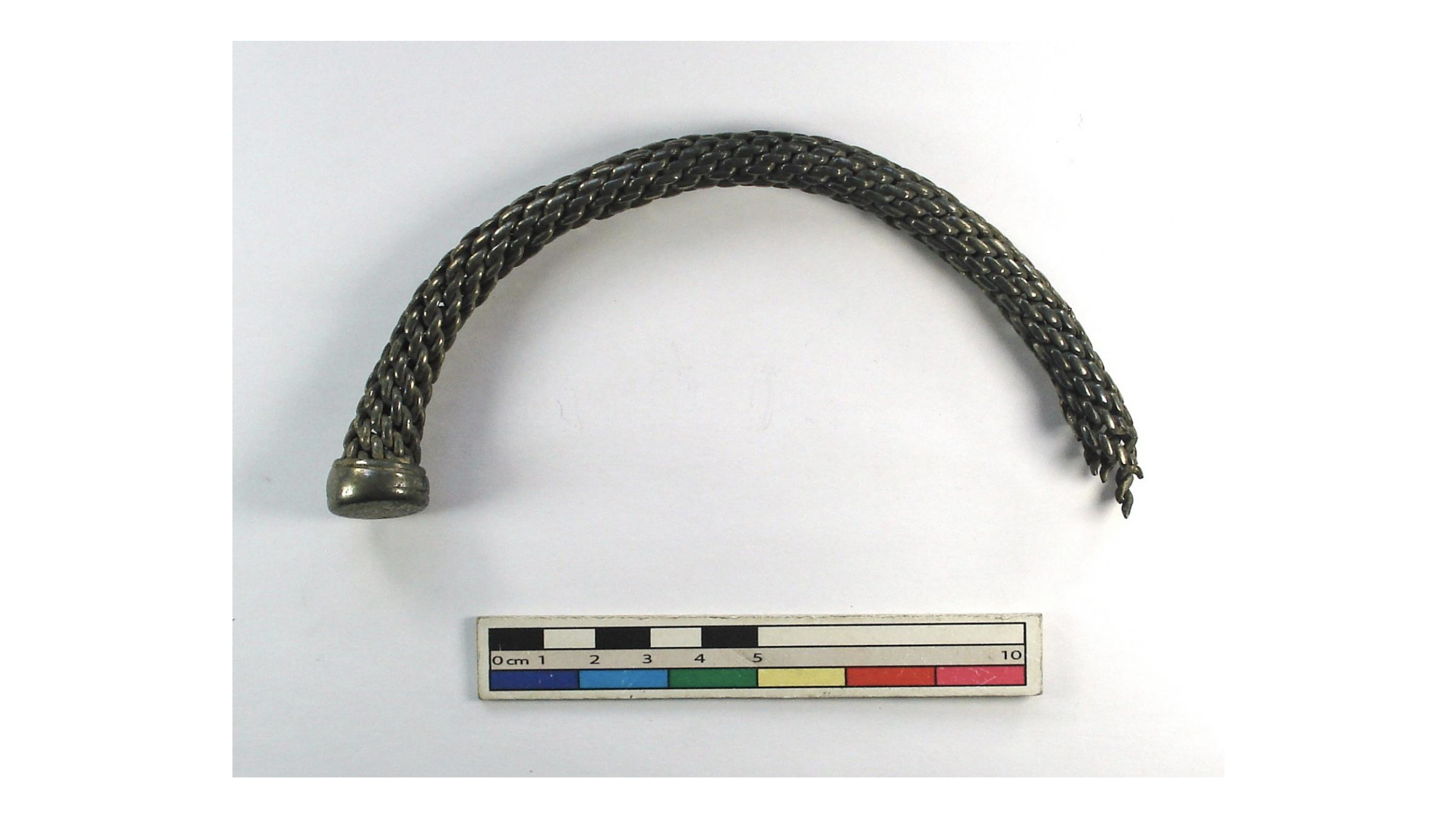} & \includegraphics[width=.145\textwidth,trim={0cm 2cm 0cm 3cm},clip]{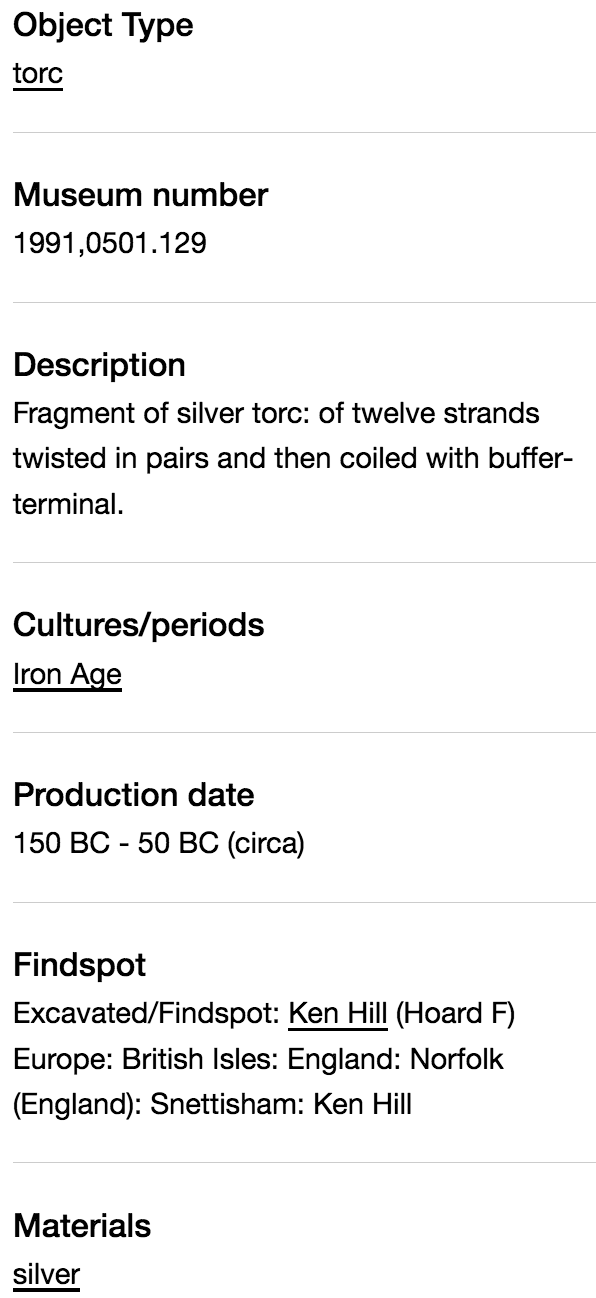} \\
        \end{tabular}
    \caption{A torc fragment from the British Museum and some of the information provided about it in the museum's collection catalogue$^{\ref{footnote:torc}}$ (image © The Trustees of the British Museum).}
    \label{fig:torc}
\end{figure}

The first example involves physical measurement where general scientific reproducibility definitions apply entirely uncontroversially. The torc (or neck ring) shown in Figure~\ref{fig:torc} is from the British Museum collection.\footnote{\label{footnote:torc}\url{https://www.britishmuseum.org/collection/object/H_1991-0501-129}} The information provided about it in the collection catalogue (some of it also shown in Figure~\ref{fig:torc}) includes a measurement of its weight which is given as 87.20g.  


 
%

Museum records contain two other weight measurements of torc 1991,0501.129, and museum staff performed four additional weighings for this study, yielding a set of seven measured quantity values for the mass of this torc, shown in the last column of Table~\ref{tab:torc-conds}. Details such as the scales used, whether they were calibrated, placed on a flat surface, etc., are not normally recorded in a museuym context, but the general expectation is nevertheless that the same readings are obtained. The values in Table~\ref{tab:torc-conds} range from 87.2g to 92g, a sizable difference that implies
there must have been some differences in the conditions under which the 
measurements were performed that resulted in the different measured values. 

Museum records show\footnote{Information about museum records and practices in this and other sections very kindly provided by Dr Julia Farley, curator of the British Museum's European Iron Age and Roman Conquest Period collections.} that the measurements were taken by different people (\textit{Team}) at different times, up to 30 years apart (\textit{Date}); 
staff normally place the object on ordinary electronic scales noting down the displayed number and rounding to one tenth of a gramme (\textit{Measurement method}); calibration is sometimes checked with a 10g standard weight, but scales are not recalibrated  (\textit{Measurement procedure}). 
The museum catalogue also records details of a conservation treatment in 1991 for torc 1991,0501.129 (for full details see the link in Footnote~\ref{footnote:torc}) which included dirt removal with a scalpel, and immersion in acid (\textit{Object conditions}). 

Table~\ref{tab:torc-conds} provides an overview of the conditions of measurement mentioned above (\textit{Team, Date}, etc.), alongside the corresponding values for condition in each of the seven measurement where available. Despite some missing information, the information indicates that the treatment for dirt and corrosion removal in 1991 led to the biggest reduction in weight, with the exact scales and use of a standard weight potentially explaining some of the remaining differences.

\definecolor{LightAzul}{rgb}{0.9,0.94,1}
\newcolumntype{a}{>{\columncolor{LightAzul}}c}

\begin{table*}[]
    \centering
\setlength\tabcolsep{3pt} 
\renewcommand{\arraystretch}{1.25} 
\sffamily\selectfont
\begin{scriptsize}\begin{tabular}{|l|c||a|a|a||l|c||r|}
    \hline
    \multicolumn{1}{|c|}{} & \multicolumn{1}{c||}{}& \textit{Object} & {\textit{Measurement}} & \multicolumn{1}{a||}{\textit{Measurement pro-}} & &  & \multicolumn{1}{c|}{Measured} \\
    \multicolumn{1}{|c|}{Object} & \multicolumn{1}{c||}{Measurand}& \textit{conditions} & {\textit{method conditions}} & \multicolumn{1}{a||}{\textit{cedure conditions}} & Date& Team & \multicolumn{1}{c|}{quantity} \\
      \hhline{~~|-|-|-||~~~}
    \multicolumn{1}{|c|}{} &  & \textit{Treatments} & \multicolumn{1}{a|}{\textit{Scales}} & {\textit{Standard weight}}&  &   & \multicolumn{1}{c|}{value} \\
    \hline
    \hline
    {1991,0501.129} & mass & \textit{0} & \textit{?} & \textit{?} & 1991 & ? &  92 g \\
    \hline
    {1991,0501.129} & mass & \textit{0?} & \textit{?} & \textit{?} & ? & JF? &  92.0 g \\ 
    \hline
    {1991,0501.129} & mass & \textit{1} & \textit{?} & \textit{10g} & 2012 & JF&  87.2 g \\ 
    \hline
    {1991,0501.129} & mass & \textit{1} & \textit{SWS pocket scales} & \textit{none} & 2021 & CM & 87.47 g \\ 
    \hline
    {1991,0501.129} & mass & \textit{1} & \textit{SWS pocket scales} & \textit{none} & 2021 & CM & 87.37 g \\ 
    \hline
    {1991,0501.129} & mass & \textit{1} & \textit{CBD bench counting scales} & \textit{none} & 2021 & CM & 88.1 g \\ 
    \hline
    {1991,0501.129} & mass & \textit{1} & \textit{CBD bench counting scales} & \textit{none} & 2021 & CM & 88.1 g \\ 
    \hline
\end{tabular}\end{scriptsize}
    \caption{Some of the conditions of measurement for the seven weight measurements of torc 1991,0501.129. SWS = Smart Weigh Digital Pocket scale SWS100; CBD = Adam CBD 4 bench counting scales.}
    \label{tab:torc-conds} 
\end{table*}

\subsection{Weighted F1-score of a text classifier}\label{ssec:f1}

The second example involves non-physical measurements in the form of a set of eight weighted F1 (wF1) scores for the same NLP system variant of which seven were obtained in four reproduction studies of \citet{vajjala-rama-2018-experiments}. In the original paper, \citeauthor{vajjala-rama-2018-experiments} report variants of a text classifier that assigns grades to essays written by second-language learners of German, Italian and Czech. One of the multilingual system variants, referred to below as multPOS$^{-}$, uses part-of-speech (POS) n-grams, and doesn't use language identity information. The information provided about it in the paper includes a wF1 score of 0.726 (Table~3 in the paper). 

We have seven other wF1 scores for multPOS$^{-}$ from subsequent research which intended, in at least the first experiment in each case, to repeat the original experiment exactly.
\citet{arhiliuc-etal-2020-language} report a score of 0.680 wF1 (Table~3 in the paper). \citet{huber-coltekin-2020-reproduction} report 0.681 (Table~3 in the paper). \citet{bestgen-2020-reproducing} reports three wF1 scores: 0.680 and 0.722 (Table~2 in the paper) and 0.728 (Table~5). \citet{caines-buttery-2020-reprolang} report 2 scores: 0.680 and 0.732 (Table~4 in the paper). Four of the measured quantity values (those shown first for each team in Table~\ref{tab:f1-conds}) were obtained in conditions as close to identical as the teams could manage, but nevertheless range from 0.680 to 0.726, a difference in wF1 score that would be considered an impressive improvement in many NLP contexts. 

From the papers we know that the scores were produced by five different teams (\textit{Team}), up to two years apart (\textit{Date}). wF1 measurements are normally performed by computing the wF1 score over paired system and target outputs, but no information about implementation of the wF1 algorithm (\textit{Measurement method}) is given. In most cases, the original system code by Vajjala \& Rama (V\&R) was used, but Caines \& Buttery (C\&B) reimplemented it, and Bestgen (B) and Huber \& \c{C}{\"o}ltekin (H\&C) produced cross-validated results with 10 random seeds (\textit{Object conditions}), rather than a single fixed seed. The measured values were all obtained in different compile-time (CT) (\textit{Object conditions}), and run-time (RT) environments  (\textit{Measurement procedure}), except for two results where a Docker container was used. The test set was the same as the original in all but three cases (marked as B, C\&B and H\&C in the Inputs column) (\textit{Measurement procedure}).

Table~\ref{tab:f1-conds} provides an overview of the conditions mentioned above (\textit{Team, Date}, etc.) under which the eight measurements were obtained. Conditions can be seen as attribute-value pairs where the attribute name is here shown in the (lower) column heading, and the values for it in the table cells. 
We know the reason for some of the differences, because the authors controlled for them: e.g.\ the differences in compile time and run time environments explain the difference between Bestgen's first and second results, and reimplementation of the system code in R explains the difference between Caines \& Buttery's first and second results. Despite missing information, we can see that different CT/RT, and performing 10-fold cross-validation vs.\ a single run with fixed seed, account for some of the remaining differences. 

\begin{table*}[]
    \centering
\setlength\tabcolsep{2.75pt} 
\renewcommand{\arraystretch}{1.25} 
\sffamily\selectfont
\begin{scriptsize}\begin{tabular}{|c|c||a|a|a|a|a|a|a|a||c|c||c|}
    \hline
     & & \multicolumn{3}{a|}{} & \multicolumn{2}{a|}{{\textit{Meas.\ method}}} & \multicolumn{3}{a||}{{\textit{Measurement pro-}}} & & & \multicolumn{1}{c|}{{Measured}} \\
    Object & Measurand & \multicolumn{3}{a|}{{\textit{Object conditions}}} & \multicolumn{2}{a|}{{\textit{conditions}}} & \multicolumn{3}{a||}{{\textit{cedure conditions}}} & Date & Team & \multicolumn{1}{c|}{{quantity}} \\
      \hhline{~~--------~~~}
    \multicolumn{1}{|c|}{} &  & {\textit{Code*}} & {\textit{Seed}} & {\textit{CT env}} & \textit{Method} & \multicolumn{1}{a|}{{\textit{Implem.\ }}} & \textit{Procedure} & {\textit{Inputs}} & \multicolumn{1}{a||}{{\textit{RT Env.\ }}} & & & {value} \\
    \hline
    \hline
    \scriptsize{multPOS$^-$}& wF1  & {\textit{V\&R}}\tablefootnote{\url{https://github.com/nishkalavallabhi/UniversalCEFRScoring}} & {\textit{V\&R 1 fixed}} & {\textit{V\&R}} & \textit{wF1(o,t)} & {\textit{V\&R}} & \textit{OTE} & {\textit{V\&R}} & {\textit{V\&R}} & 2018 & {{V\&R}} &  0.726 wF1 \\ 
    \hline 
    \scriptsize{multPOS$^-$}& wF1 & {\textit{V\&R}} & {\textit{?}} & {\textit{A et al.\ Win}} & \textit{wF1(o,t)} & {\textit{V\&R?}} & \textit{OTE} & {\textit{V\&R}} & {\textit{A et al.\ Win}} & 2020 & {{A et al.}} &  0.680 wF1 \\ 
    \hline
    \scriptsize{multPOS$^-$}& wF1     & {\textit{V\&R}} & {\textit{V\&R 1 fixed}} & {\textit{B MacOS}} & \textit{wF1(o,t)} & {\textit{?}} & \textit{OTE} & {\textit{V\&R}} & {\textit{B MacOS}} & 2020 & {{B}} &  0.680 wF1 \\
    \hline
    \scriptsize{multPOS$^-$}& wF1    & {\textit{V\&R}} & {\textit{V\&R 1 fixed}} & {\textit{B Docker}} & \textit{wF1(o,t)} & {\textit{?}} & \textit{OTE} & {\textit{V\&R}} & {\textit{B Docker}} & 2020 & {{B}} &  0.722 wF1 \\
    \hline
    \scriptsize{multPOS$^-$}& wF1     & {\textit{V\&R+s}} & {\textit{B 10 avg}} & {\textit{B Docker}} & \textit{wF1(o,t)} & {\textit{?}} & \textit{OTE} & {\textit{B}} & {\textit{B Docker}} & 2020 & {{B}} &  0.728 wF1 \\
    \hline
    \scriptsize{multPOS$^-$}& wF1  & {\textit{V\&R}} & {\textit{V\&R 1 fixed}} & {\textit{C\&B1}} & \textit{wF1(o,t)} & {\textit{?}} & \textit{OTE} & {\textit{V\&R}} & {\textit{C\&B1}} & 2019 & {{C\&B}} & 0.680 wF1 \\ 
    \hline
    \scriptsize{multPOS$^-$}& wF1  & {\textit{C\&B}}\tablefootnote{\url{https://github.com/cainesap/CEFRgrader}} & {\textit{?}} & {\textit{C\&B2}} & \textit{wF1(o,t)} & {\textit{?}} & \textit{OTE} & {\textit{C\&B}} & {\textit{C\&B2}} & 2020 & {{C\&B}} & 0.732 wF1 \\ 
    \hline
    \scriptsize{multPOS$^-$}& wF1  & {\textit{V\&R}} & {\textit{H\&C 10 avg}} & {\textit{H\&C}} & \textit{wF1(o,t)} & {\textit{?}} & \textit{OTE} & {\textit{H\&C}} & {\textit{H\&C}} & 2020 & {{H\&C}} &  0.681 wF1 \\
    \hline
\end{tabular}\end{scriptsize}
    \caption{Some conditions of measurement for eight weighted F1 measurements of Vajjala \& Rama's multilingual POS-ngram CEFR classifier system without language category information variant (multPOS$^-$). *Code here shown separately from random seed. OTE = outputs vs.\ targets evaluation, i.e.\ the standard procedure of obtaining outputs for a set of test inputs and computing metrics over system and target outputs.}
    \label{tab:f1-conds} 
\end{table*}

\spewfootnotes

\subsection{Comparison}

In presenting and discussing the physical and non-physical measurements and conditions in the preceding two sections, we introduced common terminology for which full VIM definitions will be provided in the next section. The intention in using the same terms and layout 
in Tables~\ref{tab:torc-conds} and~\ref{tab:f1-conds} was to bring out the similarities between the two sets of measurements,  
and to show that measurements can generally, whether physical or not, be characterised 
in the same way on the basis of the same general scientific terms and definitions.

The two fields, curating of museum artifacts and NLP/ML, also have in common that (i) conditions of measurement are not traditionally recorded in any great detail, making exact repetition of a measurement virtually impossible, (ii) despite this, there is nevertheless an expectation that repeat measurements should yield the same results, and (iii) historically, practitioners have not cared about these issues very much.

\section{VIM Definitions of Repeatability and Reproducibility}\label{sec:vim-def}

The International Vocabulary of Metrology (VIM) \cite{jcgm2012international} defines repeatability and reproducibility as follows (all defined terms in boldface, see Appendix for full set of verbatim VIM definitions including subsidiary defined terms):

\begin{enumerate}[topsep=4pt,itemsep=0pt,parsep=3pt,partopsep=0pt]
 \item[{2.21}] \textbf{measurement repeatability} (or repeatability, for short) is  \textbf{measurement  precision}  under  a  set  of  \textbf{repeatability conditions of measurement}.   

\item[{2.20}] a \textbf{repeatability condition of measurement} (repeatability condition) is a condition  of  \textbf{measurement},  out  of  a  set  of  conditions   that   includes   the   same   \textbf{measurement procedure},   same   operators,   same   \textbf{measuring system},   same   operating   conditions   and   same   location,  and  replicate  measurements  on  the  same  or similar objects over a short period of time. 

\item[{2.25}] \textbf{measurement reproducibility} (reproducibility) is \textbf{measurement   precision}   under   \textbf{reproducibility conditions of measurement}. 

\item[{2.24}] a \textbf{reproducibility condition of measurement} (reproducibility condition) is a condition  of  \textbf{measurement},  out  of  a  set  of  conditions  that  includes  different  locations,  operators,  \textbf{measuring  systems}, etc. 
A   specification   should   give   the   conditions   changed and unchanged, to the extent practical.
\end{enumerate} 

\noindent The VIM definitions are squarely focused on measurement: repeatability and reproducibility are properties of measurements (not objects, scores, results or conclusions), and are defined as measurement precision, i.e.\ both are quantified by calculating the precision of a set of measured quantity values. Moreover, both concepts are defined relative to a set of conditions of measurement, in other words, the conditions have to be known and specified for assessment of either concept (repeatability, reproducibility) to be meaningful.

\section{Mapping the VIM Definitions to a Metrological Framework for NLP/ML}\label{sec:mapping}

The VIM definitions provide the definitional basis for reproducibility assessment. However, for a framework that tells us what to do in practice in an NLP/ML context, we also need (i) a method for computing precision, (ii) a specification of the practical steps in performing assessments, and (iii) a specific set of repeatability/reproducibility conditions of measurement. In this section, we present the fixed, discipline-independent elements of the proposed framework (Section~\ref{ssec:fixed-elements}),  methods for computing precision in NLP/ML (Section~\ref{ssec:cv}), a series of steps for practical assessment of reproducibility (Section~\ref{ssec:steps}), and, utilising previous work on reproducibility in the NLP/ML field,  starting points for identifying suitable conditions of measurement (Section~\ref{ssec:conditions}).

\subsection{Fixed elements of the framework}\label{ssec:fixed-elements}

\begin{figure}[t!]
    \centering
    \includegraphics[width=.475\textwidth]{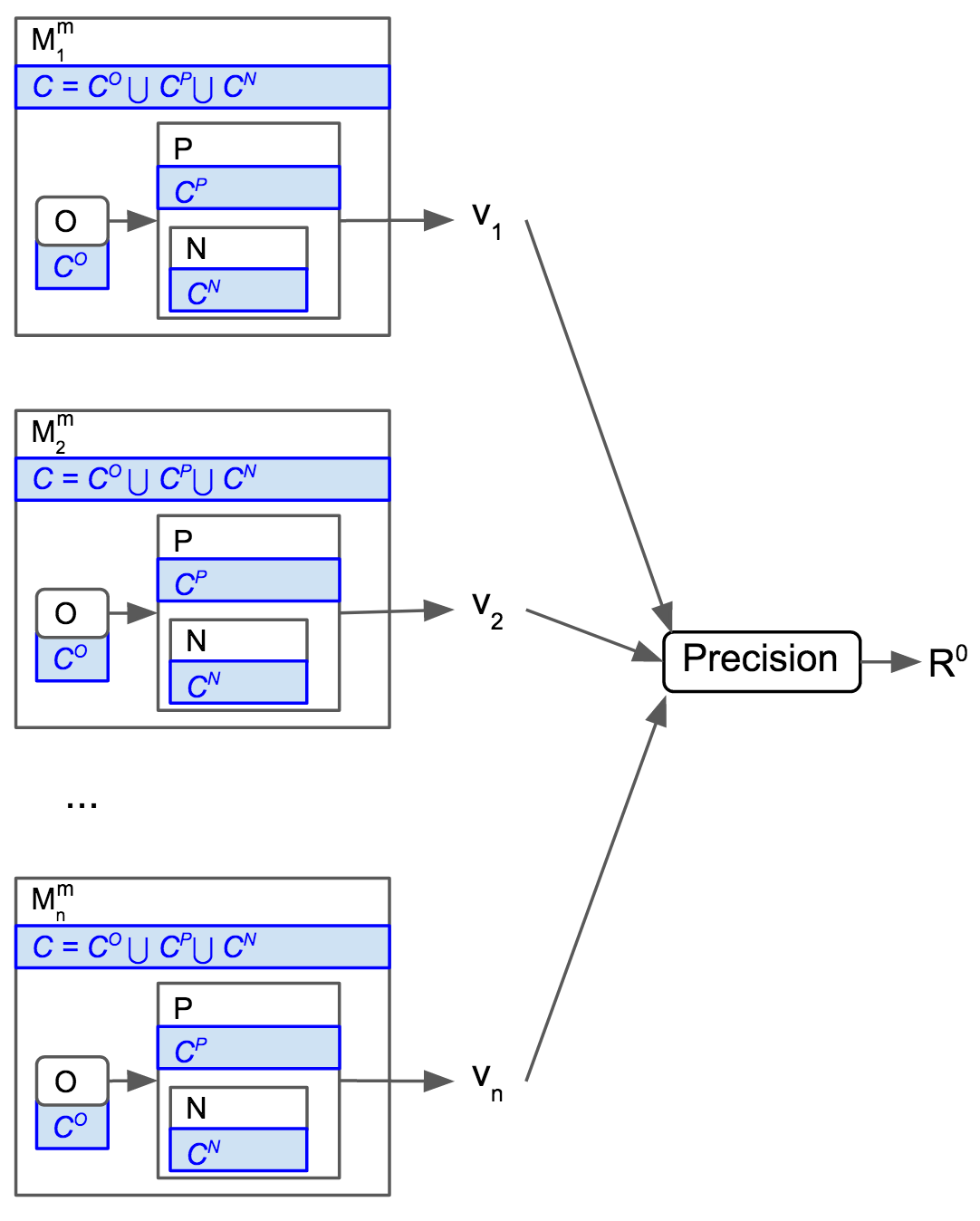}
    \caption{Diagrammatic overview of repeatability assessment of measurements $M_1, M_2, ...$ $M_n$ of object $O$, with measurand $m$, and repeatability conditions of measurements $C$ (same for all $M_i$). Repeatability $R^0$ is defined as the precision of the set of values $v_i$ returned by $M_i$. (For $C^O$, $C^N$, and $C^P$, see in text.)}
    \label{fig:diagram-repeat}
\end{figure}

Table~\ref{tab:f1-conds} gave a first indication what (some) conditions might look like in NLP/ML, grouped into \textit{Object conditions} $C^O$, \textit{Measurement method conditions} $C^N$, and \textit{Measurement procedure conditions} $C^P$. These conceptually useful groupings are retained to yield a skeleton framework shown in diagrammatic form in Figure~\ref{fig:diagram-repeat}, which corresponds to the following definition of repeatability $R^0$:

\vspace{-.4cm}
\begin{equation*}
\begin{aligned}
&R^0(M_1, M_2, ... M_n) := \,\text{Precision}(v_1, v_2, ... v_n)\\ 
&\text{where } M_i\!: (m, O, t_i, C) \mapsto v_i\\ 
\end{aligned}\label{def:repro}
\end{equation*}

\noindent and the $M_i$ are repeat measurements for measurand $m$ performed on object $O$ at different times $t_i$ under (the same) set of conditions $C = C^O \cup C^N \cup C^P$. Below the coefficient of variation is used as the precision measure, but other measures are possible. The members of each set of conditions are attribute/value pairs each consisting of a name and a value. Reproducibility $R$ is defined in the same way except that condition values differ for at least one condition in the $M_i$:

\vspace{-.4cm}
\begin{equation*}
\begin{aligned}
&R(M_1, M_2, ... M_n) := \,\text{Precision}(v_1, v_2, ... v_n)\\ 
&\text{where } M_i\!: (m, O, t_i, C_i) \mapsto v_i\\ 
\end{aligned}\label{def:repro}
\end{equation*}


\subsection{Computing precision}\label{ssec:cv}

Precision in metrological studies is reported in terms of some or all of the following: mean, standard deviation with 95\% confidence intervals, coefficient of variation, percentage of measured quantity values within $n$ standard deviations. 

In reproducibility assessment in NLP/ML, sample sizes tend to be very small (a sample size of 8 as in Section~\ref{ssec:f1} is currently unique). We therefore need to use de-biased sample estimators: we use the unbiased sample standard deviation, denoted $s^*$, with confidence intervals calculated using a t-distribution, and standard error (of the unbiased sample standard deviation) approximated on the basis of the standard error of the unbiased sample variance $\text{se}(s^2)$ as $\text{se}_{s^2}(s^*) \approx \frac{1}{2\sigma}\text{se}(s^2)$ \cite{rao1973linear}. Assuming measured quantity values are normally distributed, we calculate the standard error of the sample variance in the usual way: $\text{se}(s^2) = \sqrt{\frac{2\sigma^4}{n-1}}$. Finally, we also use a small sample correction for the coefficient of variation: $\text{CV}^* = (1+\frac{1}{4n})\text{CV}$ \cite{sokal:rohlf:1969}.

Equipped with the above, the reproducibility of wF1 measurements of \citet{vajjala-rama-2018-experiments}'s MultPOS$^-$ system can, for example, be quantified based on the eight replicate measurements from Table~\ref{tab:f1-conds} (disregarding the role of measurement conditions for the moment) as being CV$^*$ = 3.818.\footnote{Code for computing CV$^*$ available here: \url{https://github.com/asbelz/coeff-var}} This and two other example applications of the framework is presented in more detail in Section~\ref{sec:ex-appl}.

\subsection{Steps in reproducibility assessment}\label{ssec:steps} 

\begin{figure}[h!]
    \centering
\renewcommand {\footnotesize} {\fontsize {10pt} {12pt}\selectfont}
\begin{footnotesize}
\begin{tabular}{|p{.45\textwidth}|}
\hline 
\\
{\fontsize{9.9}{11.5}\textbf{2-PHASE REPRODUCIBILITY ASSESSMENT}}\\
\begin{enumerate}[topsep=0pt,itemsep=0pt,parsep=3pt,partopsep=0pt]
    \item[] \hspace{-.85cm} {\fontsize{9.9}{11.5}\textit{REPEATABILITY PHASE}}
    \item Select measurement to be assessed, identify shared object and measurand.
    \item Select initial set of repeatability conditions of measurement $C^0$ and specify value for each condition.
    \item Perform $n$$\geq$$2$ reproduction measurements to yield measured quantity values $v^0_1, v^0_2, ... v^0_n$.
    \item Compute precision for $v^0_1, v^0_2, ... v^0_n$, giving repeatability score $R^0$.
    \item Unless precision is as small as desired, identify additional conditions that had different values in some of the reproduction measurement, and add them to the set of measurement conditions, also updating the measurements to ensure same values for the new conditions. Repeat Steps~3--5.
    \vspace{.1cm}
    \item[] \hspace{-.85cm} {\fontsize{9.9}{11.5}\textit{REPRODUCIBILITY PHASE}}
    \item From the final set of repeatability conditions, select the conditions to vary, and specify the different values to test.
    \item For each combination of differing condition values:
    \begin{enumerate}    
        \item Carry out $n$ reproduction tests, yielding measured quantity values $v_1, v_2, ... v_n$
        \item Compute precision for $v_1, v_2, ... v_n$, giving reproducibility score $R$.
    \end{enumerate}
\end{enumerate} \\
Report all resulting $R$ scores, alongside baseline $R^0$ score.\vspace{.2cm}\\
\hline
\end{tabular}
\end{footnotesize}
    \caption{Steps in 2-phase reproducibility assessment with baseline repeatability assessment. Step~5 is obsolete if a field has standard conditions of measurement.}
    \label{fig:7-steps}
\end{figure}

In order to relate multiple repeatability and reproducibility assessments of the same object and measurand to each other and compare them, they need to share the same conditions of measurement (same in terms of attribute names, not necessarily values, see next section). Repeatability 
is simply the special case of reproducibility where all condition values are also the same.

For a true estimate of the variation resulting from given differences in condition values, the baseline variation, present when all condition values are the same, needs to be known. It is therefore desirable to carry out repeatability assessment prior to reproducibility assessment in order to be able to take into account the baseline amount of variation. E.g.\ if the coefficient of variation is $x_{C^0}$ under identical conditions $C^0$, and $x_{C}$ 
under varied conditions $C$, then it's the difference between $x_{C^0}$ and $x_{C}$
that estimates the effect of varying the conditions. Finally, for very small sample sizes, both baseline repeatability assessment, and subsequent reproducibility assessment, should use the same sample size to ensure accurate assessment of variation due to the varied condition values (alone).

Figure~\ref{fig:7-steps} translates what we have said in this and preceding sections into a 2-phase framework for reproducibility assessment. If a field develops shared standard conditions of measurement, Step~5 is obsolete. In situations where reproduction studies have been carried out without a pre-defined, shared set of conditions of measurement, reproducibility assessment can still be carried out (as we did for the Vajjala \& Rama system in Section~\ref{ssec:f1}), but in this situation a baseline assessment of variation under repeatability conditions of measurement is not possible, and the only option is to use a single-phase version of the framework as shown in Figure~\ref{fig:1-phase}. 
See Section~\ref{sec:disc} for more discussion of this issue.

\subsection{Conditions of measurement}\label{ssec:conditions}

The final component needed for a metrological framework for reproducibility assessment in NLP/ML is a specific set of conditions of measurement. As mentioned in Section~\ref{ssec:fixed-elements}, individual conditions consist of name and value, 
and their role is to capture those attributes of a measurement where different values may cause differences in measured values.
As indicated by Step~5 in Figure~\ref{fig:7-steps}, for repeatability, conditions should be selected with a view to reducing baseline variation. 
As it's not practically feasible (or, normally, theoretically possible) to specify and control for all conditions that can be the same or different in a measurement, some judgment is called for here (see Section~\ref{sec:disc} for discussion). The  idea is that a discipline evolves shared standard conditions that address this.

It so happens that much of the reproducibility work in ML and NLP has so far focused on what standard conditions of measurement (information about system, data, dependencies, computing environment, etc.)\ for \textit{metric} measurements need to be specified in order to enable repeatability assessment, even if it hasn't been couched in these terms.  Out of all reproducibility topics, pinning down the information and resources that need to be shared to enable others to obtain the same metric scores is the one that has attracted by far the most discussion and papers. In fact, reproduction in NLP/ML has become synonymous with rerunning code and obtaining the same metric scores again as were obtained in a previous study. 

Reproducibility checklists such as those provided by \cite{pineau2020checklist} and the ACL\footnote{\url{https://2021.aclweb.org/calls/reproducibility-checklist/}} are lists of types of information (attributes) for which authors are asked to provide information (values), and these  can directly be construed as conditions of measurement. 
In this section, the intention is not to propose definitive sets of conditions relating to object, measurement method and measurement procedure that should be used in NLP/ML. Rather, in each subsection, we point to existing research such as the above that can be used to provide a `starter set' of conditions of measurement.

\begin{figure}[t]
    \centering
\renewcommand {\footnotesize} {\fontsize {10pt} {12pt}\selectfont}
\begin{footnotesize}\begin{tabular}{|p{.45\textwidth}|}
\hline 
\\
{\fontsize{9.9}{11.5}\textbf{1-PHASE REPRODUCIBILITY ASSESSMENT}}\\
\begin{enumerate}[topsep=0pt,itemsep=0pt,parsep=3pt,partopsep=0pt]
    \item For set of $n$ measurements to be assessed, identify shared object and measurand.
    \item Identify all conditions of measurement $C$ for which information is available for all measurements, and specify values for each condition.
    \item Gather the $n$ measured quantity values  $v_1, v_2, ... v_n$
    \item Compute precision for $v_1, v_2, ... v_n$, giving reproducibility score $R$.
\end{enumerate} \\
Report  resulting $R$ score.\vspace{.2cm}\\
\hline
\end{tabular}
\end{footnotesize}
    \caption{Steps in 1-phase reproducibility assessment for assessing a set of existing measurements where baseline repeatibility assessment is not possible. See also discussion in Section~\ref{sec:disc}.}
    \label{fig:1-phase}
\end{figure}

\subsubsection{Object conditions}

The ML Code Completeness Checklist\footnote{\url{https://github.com/paperswithcode/releasing-research-code}} (adopted as part of the NeurIPS'21 guidelines) consists of five items: specification of dependencies,
training code, evaluation code, pre-trained models, README file including results and commands. These
provide a good starting point for object conditions of measurement 
(note we group evaluation code under measurement method conditions, see next section):

\vspace{.15cm}
\begin{enumerate}[topsep=0pt,itemsep=0pt,parsep=2pt,partopsep=0pt]
    \item Dependencies
    \item Training code
    \item Pre-trained models
    \item Precise commands to run code/produce results
    \item Compile-time environment
    \item Run-time environment
\end{enumerate}
\vspace{.15cm}

\noindent In human evaluation of system outputs, object conditions don’t apply, as the object of measurement is fully specified by a sample of its outputs.

\subsubsection{Measurement method conditions}

In metric-based measurement where the measurand is defined mathematically, a measurement method is an implementation of a method for (or, conceivably, a manual way of) computing a quantity value for the measurand. In 
this case, the method, like the Object of measurement, is a computational artifact, so the same conditions can be used as in the preceding section (albeit with different names to mark the difference; values will clearly also differ). 

In human-evaluation-based measurement, a different set of conditions is needed. Here, the measurand is identified by (ideally standardised) name and definition for the quantity being assessed. This has been termed {`quality criterion'}\footnote{In this and subsequent sections we use five closely related terms which relate to each other as follows \cite{belz-etal-2020-disentangling}: (i) quality criterion + evaluation mode = evaluation measure; (ii) evaluation measure + experimental design = evaluation method.} in previous work \cite{howcroft-etal-2020-twenty,belz-etal-2020-disentangling} which took a first stab at a set of standardised names and definitions. These, in conjunction with a checklist for human evaluation such as the one proposed by \citet{shimorina2021human}, can provide a starting point for measurement method conditions (as follows), and measurement procedure conditions (next section), for human evaluation. 

\vspace{.15cm}
\begin{enumerate}[topsep=0pt,itemsep=0pt,parsep=2pt,partopsep=0pt]
    \item Name and definition of measurand 
    \item Evaluation mode\footnote{Absolute vs.\ relative, intrinsic vs.\ extrinsic, objective vs.\ subjective.}
    \item Method of response elicitation
    \item Method for aggregating or otherwise processing  raw  participant responses
    \item Any code used (conditions as for Object)
\end{enumerate}

\subsubsection{Measurement procedure conditions}

Measurement procedure conditions capture any information needed to apply a given measurement method in practice. In metric-based measurement, this includes:

\vspace{.15cm}
\begin{enumerate}[topsep=2pt,itemsep=0pt,parsep=3pt,partopsep=0pt]
    \item Test set
    \item Any preparatory steps taken, such as preprocessing of text
    \item Any code used (conditions as for Object)
\end{enumerate}
\vspace{.15cm}

\noindent In human-evaluation-based measurement, some of the remaining properties from \citet{shimorina2021human} and \citet{howcroft-etal-2020-twenty} can be used:

\vspace{.15cm}
\begin{enumerate}[topsep=2pt,itemsep=0pt,parsep=3pt,partopsep=0pt]
        \item If test set evaluation, test set  and  preprocessing code/method(s); if system interaction, specification of system set up
        \item Responses  collection method 
        \item Quality  assurance code/method(s) 
        \item Instructions to evaluators 
        \item Evaluation interface 
            \item Any code used (conditions as for Object)

\end{enumerate}

\section{Examples}\label{sec:ex-appl}

In this section we apply the framework to the two sets of measurements from Sections~\ref{ssec:torc} and~\ref{ssec:f1}, and an additional set of measurements from a recent reproduction study \cite{mille:etal:2021}.

\subsection{Mass of a Torc}

In the case of the torc mass measurements, the reproducibility analysis is performed post hoc, i.e.\ we cannot obtain more information for the three older measurements than was recorded at the time. We therefore have to use the 1-phase assessment (Figure~\ref{fig:1-phase}), and for the complete set of seven weighings specify all conditions (Table~\ref{tab:torc-conds}) as different. CV$^*$ results can then be reported as follows:

\vspace{.1cm}

\hspace{-.2cm}\begin{tabular}{p{.45\textwidth}}
    \textit{Mass measurement reproducibility under reproducibility conditions of measurement as detailed in Table~\ref{tab:torc-conds}, was assessed on the basis of seven measurements of torc 1991,0501.129 as follows: the unbiased coefficient of variation is \textbf{2.61},
for a mean of 88.89, 
unbiased sample standard deviation of 2.24 with 95\% CI (0.784, 3.696),
and sample size 7.
All measured values fall within two standard deviations, 71.43\% within one standard deviation. }
\end{tabular}
\vspace{.01cm}

\noindent If we used just the four measurements for which all condition values are known, we would get CV$^*$ = 0.519 under reproducibility conditions where only the scales used differ. For a sample size of 4 we can still be reasonably confident that this is a good estimate of CV$^*$ for the whole population (stdev 95\% CI [-0.04, 0.90]). 

\textit{Repeatability} assessment can be performed for the two subsets of measurements for the SDS scales and the CBD scales, which gives CV$^*$ = 0.11 for the former and CV$^*$ = 0 for the latter, for similar-size stdev confidence intervals.

\subsection{wF1 of a Text Classifier}

The situation is similar for the text classifier wF1 measurements (Table~\ref{tab:f1-conds}) in that we are restricted to the information made available by the authors, which is however, more complete than in the torc example. Here too we have to use the 1-phase assessment (Figure~\ref{fig:1-phase}), and for the complete set of 8 wF1 values specify all conditions as different. CV$^*$ results can then be summed up as above:

\vspace{.1cm}

\hspace{-.2cm}\begin{tabular}{p{.45\textwidth}}
    \textit{wF1 measurement reproducibility under reproducibility conditions of measurement as in Table~\ref{tab:f1-conds}, 
    was assessed on the basis of eight measurements reported by \citet{vajjala-rama-2018-experiments}, \citet{arhiliuc-etal-2020-language}, \citet{bestgen-2020-reproducing}, \citet{caines-buttery-2020-reprolang}, and \citet{huber-coltekin-2020-reproduction} (Table~\ref{tab:f1-conds}): CV$^*$ = 
    \textbf{3.818}, mean = 0.7036, unbiased sample standard deviation = 0.0261, 95\% CI [0.01, 0.04], sample size = 8. All measured values$\,$fall within$\,$two $\!$ standard $\!$ deviations,} \\
\end{tabular}\vspace{.2cm}
\begin{tabular}{p{.45\textwidth}}
     \textit{87\% within one standard deviation.}\\
\end{tabular}

\subsection{Clarity and Fluency of an NLG system}

The third example comes from a recent reproduction study \cite{mille:etal:2021} which repeated the human evaluation of a Dutch-language football report generation system \cite{van2017pass}. There were two main measurands, mean Clarity ratings and mean Fluency ratings, and a single object (the report generator) was evaluated. There were two scores for each of the measurands, one from the original study, one from the reproduction study. Here the situation was that a repeatability study was intended, but not possible.\footnote{Original evaluators could not be used, and COVID-19 pandemic restrictions prevented real-world interaction.} Therefore a different set of evaluators and a different evaluation interface had to be used.

Both Clarity and Fluency ratings were obtained on a 7-point scale (1..7). Computed on the scores as reported, for Clarity, CV$^* = 10.983$, for Fluency, CV$^* = 13.525$. However, if the 7-point scale had been 0..6, higher CV$^*$ values would have been obtained, i.e.\ results are not comparable across different rating scales. To address this, rating scales should be mapped to range with lowest score 0. Results can then be reported as follows:

\vspace{.1cm}

\hspace{-.2cm}\begin{tabular}{p{.45\textwidth}}
    \textit{Clarity measurement reproducibility under reproducibility conditions of measurement as detailed in \citet{mille:etal:2021},  
    was assessed on the basis of 2 measurements reported by \citet{van2017pass} and \citet{mille:etal:2021}, rescaled to 0..6: CV$^*$ = \textbf{13.193}, mean = 4.969, unbiased sample standard deviation = 0.583, 95\% CI [-2.75, 3.92], sample size = 2. Measured values fall within one standard deviation.}\\
\end{tabular}
\vspace{.1cm}

\hspace{-.2cm}\begin{tabular}{p{.45\textwidth}}
    \textit{Fluency measurement reproducibility under reproducibility conditions of measurement as detailed in \citet{mille:etal:2021},  
    was assessed on the basis of 2 measurements reported by \citet{van2017pass} and \citet{mille:etal:2021}, rescaled to  0..6: CV$^*$ = \textbf{16.372}, mean = 4.75, unbiased sample standard deviation = 0.691, 95\% CI [-3.26, 4.645], sample size = 2. Measured values fall within one standard deviation.}\\
\end{tabular} 

\subsubsection{Notes on Examples}

CV$^*$ is fronted in the examples above as the `headline' result. It can take on this role, because it is a general measure, not in the unit of the measurements (unlike mean and standard deviation), providing a quantification of precision (degree of reproducibility) that is comparable across studies \cite[p.\ 57]{ahmed1995pooling}. This also holds for percentage within $n$ standard deviations but the latter is a less recognised measure, and likely to be the less intuitive for many.

CV is a measure of precision and as such, sample size should be $\geq3$. For a sample of 2 (as in the human evaluation in the last section), CV$^*$ is still meaningful as a measure of the variation found in the sample. However, it will generally provide a less reliable estimate of population CV.

\section{Discussion}\label{sec:disc}

Specifying conditions of measurement in reproducibility assessment is important so results can be compared across different measures and assessments. We have not attempted to come up with a definitive set of conditions, but pointed to other research as a starting point. One important role the conditions play is to facilitate estimation of baseline variation via repeatability testing. It could be argued that if the goal is to ensure as near as possible identical measurements in repeatability testing, then a straightforward way to achieve that is containerisation. However, firstly the purpose of repeatability testing is to assess variation under normal use and it's not realistic to always run systems in Docker containers even in a research context. Secondly, human evaluation can't be run in a container.

What counts as a good level of reproducibility can differ substantially between disciplines and contexts. E.g.\ in bio-science assays,\footnote{An investigative analytic procedure in the physical sciences e.g.\ laboratory medicine.} precision (reported as coefficient of variation) ranges from typically <10\% for enzyme assays, to 20--50\% for in vivo and cell-based assays, and >300\% for virus titer assays \cite{aahfws:nd}. For NLP, such typical CV ranges would have to be established over time, but it seems clear that we would expect typical CV for metric-based measurements to be much lower (better) than for human-assessment-based measurements. For a set of wF1 measurements, the 3.8\% CV$^*$ above seems high.

There are many ways in which results from similar studies can be compared and conclusions drawn from comparisons. For example, to make (subjective) judgments of whether the same conclusions can be drawn from a set of comparable experimental results, or to ask a group of assessors to make such judgments, is a valid and informative thing to do, but it's not reproducibility assessment in the general scientific sense of the term. It can also be informative to consider the ease with which systems can be recreated, but reproducibility is not a property of systems. 
Computer science has a history of departing from standard scientific reproducibility definitions, e.g.\ the ACM changed its definitions after NISO asked it to ``harmonize its terminology and definitions with those used in the broader scientific research community.'' \cite{acm2020artifact}. The question is, if the standard scientific terminology and definitions work for computer science, why would we not use them exactly as they are, rather than adapt them in often fundamental ways?

\section{Conclusion}\label{sec:concl}

The reproducibility debate in NLP/ML has long been framed in terms of pinning down exactly what information we need to share so that others are guaranteed to get the same metric scores (e.g.\  \citeauthor{sonnenburg2007need}, \citeyear{sonnenburg2007need}).
What is becoming clear, however, is that no matter how much of our code, data and ancillary information we share, residual amounts of variation remain that are stubbornly resistant to being eliminated. 
A recent survey \cite{belz2021systematic} found that just 14\% of the 513 original/reproduction score pairs analysed were exactly the same. Judging the remainder simply `not reproduced' would be of limited usefulness, as some are much closer to being the same than others, while assessments of whether the same conclusions can be drawn are prone to low levels of agreement. Quantifying closeness of results, and, over time, establishing expected levels closeness, seems a better way forward.

In this paper our aim has been to challenge the assumption that the general scientific reproducibility terms and definitions are not applicable or suitable for NLP/ML by directly mapping them to a practical framework that yields quantified assessments of degree of reproducibility that are comparable across different studies. 

The NLP/ML field certainly needs \textit{some} way of assessing degree of reproducibility that is comparable across studies, because the ability to assess reproducibility of results, hence the trustworthiness of evaluation practices, is a cornerstone of scientific research that the field has not yet fully achieved.

\section*{Acknowledgements}

The contribution to this work made by Dr Julia Farley, curator (European Iron Age and Roman Conquest period collections) at the British Museum, and colleagues, is gratefully acknowledged. In particular their time and patience in obtaining the seven historical and new weighings of a 2,000 year old torc, as well as providing detailed information about the weighings.

\bibliography{anthology,custom}
\bibliographystyle{acl_natbib}

\vfill
\pagebreak

\appendix
\onecolumn\section{Verbatim VIM Definitions}\label{sec:appendix}

\begin{table*}[h!]
\begin{small}
    \centering
\noindent\begin{tabular}{|p{0.3\textwidth}|p{0.635\textwidth}|}
\hline
\textbf{Primary term} (synonyms) & Definition \\
\hline
\textbf{2.21} \textbf{measurement repeatability} (repeatability) & \textbf{measurement  precision}  under  a  set  of  \textbf{repeatability conditions of measurement} \\
\hline
\textbf{2.20} (Note 1) \textbf{repeatability condition of measurement} (repeatability condition) & condition  of  \textbf{measurement},  out  of  a  set  of  conditions   that   includes   the   same   \textbf{measurement procedure},   same   operators,   same   \textbf{measuring system},   same   operating   conditions   and   same   location,  and  replicate  measurements  on  the  same  or similar objects over a short period of time \\
& NOTE  1   \hspace{.2cm}   A  condition  of  measurement  is  a  repeatability  condition   only   with   respect   to   a   specified   set   of   repeatability conditions. \\
\hline
\textbf{2.25} \textbf{measurement reproducibility} (reproducibility) & \textbf{measurement   precision}   under   \textbf{reproducibility conditions of measurement} \\
\hline
\textbf{2.24} (Note 2) \textbf{reproducibility condition of measurement} (reproducibility condition) & condition  of  \textbf{measurement},  out  of  a  set  of  conditions  that  includes  different  locations,  operators,  \textbf{measuring  systems},  and  replicate  measurements  on the same or similar objects \\
& NOTE   2 \hspace{.2cm}    A   specification   should   give   the   conditions   changed and unchanged, to the extent practical.\\
\hline
\textbf{1.1 quantity} & property  of  a  phenomenon,  body,  or  substance,  where  the  property  has  a  magnitude  that  can  be  expressed as a number and a reference\\
\hline
\textbf{1.19 quantity value} (value of a quantity, value) & number  and  reference  together  expressing  magnitude of a \textbf{quantity}\\
\hline
\textbf{2.1} \textbf{measurement} & process  of  experimentally  obtaining  one  or  more  \textbf{quantity  values}  that  can  reasonably  be  attributed  to a \textbf{quantity}\\
\hline
\textbf{2.3} \textbf{measurand} & \textbf{quantity} intended to be measured\\
\hline
\textbf{2.5 measurement method} & method of measurement generic   description   of   a   logical   organization   of   operations used in a \textbf{measurement} \\
\hline
\textbf{2.6} \textbf{measurement procedure} & detailed description of a \textbf{measurement} according to one  or  more  \textbf{measurement  principles}  and  to  a  given \textbf{measurement  method},  based  on  a  \textbf{measurement  model}  and  including  any  calculation  to  obtain a \textbf{measurement result} \\
& NOTE  1    \hspace{.2cm}   A  measurement  procedure  is  usually  documented   in   sufficient   detail to   enable   an   operator   to   perform a measurement. \\
\hline
\textbf{2.9 measurement result} (result of measurement) & set  of  \textbf{quantity  values}  being  attributed  to  a  \textbf{measurand}  together  with  any  other  available  relevant  information \\
\hline
\textbf{2.10 measured quantity value} (value of a measured quantity, measured value) & \textbf{quantity value} representing a \textbf{measurement result}\\
\hline
\textbf{2.15} \textbf{measurement precision} (precision) & closeness  of  agreement  between  \textbf{indications}  or  \textbf{measured  quantity  values}  obtained  by  replicate  \textbf{measurements}  on  the  same  or  similar  objects  under specified conditions \\
\hline
\textbf{4.1} \textbf{indication} & \textbf{quantity  value}  provided  by  a  \textbf{measuring  instrument} or a \textbf{measuring system}\\
\hline
\end{tabular}
\end{small}
    \caption{Verbatim VIM definitions of repeatability, reproducibility and related concepts \cite{jcgm2012international}. (In VIM, definitions also give   the  earlier  definition number  from  the  second  edition in parentheses which we omit here. In Definition 2.20, Note 2 relating only to chemistry is omitted. In 2.5 and 2.6, less important notes are omitted for space reasons.) \vspace{.3cm}}
    \label{tab:vim}
\end{table*}

\end{document}